\pdfoutput=1
\documentclass{article}
\PassOptionsToPackage{numbers, compress}{natbib}
\usepackage[preprint]{neurips_2026}

\usepackage{amsmath,amsfonts,bm}

\def\eqref#1{equation~\ref{#1}}

\def\1{\bm{1}}

\DeclareMathAlphabet{\mathsfit}{\encodingdefault}{\sfdefault}{m}{sl}
\SetMathAlphabet{\mathsfit}{bold}{\encodingdefault}{\sfdefault}{bx}{n}

\usepackage[utf8]{inputenc}
\usepackage[T1]{fontenc}
\usepackage{hyperref}
\usepackage{url}
\usepackage{booktabs}
\usepackage{amsfonts}
\usepackage{amsmath}
\usepackage{amssymb}
\usepackage{nicefrac}
\usepackage{microtype}
\usepackage{xcolor}
\usepackage{colortbl}
\usepackage{graphicx}
\usepackage{tabularx}
\usepackage{array}
\usepackage{placeins}
\usepackage{flafter}

\definecolor{PearlGroupGray}{gray}{0.92}
\definecolor{PearlBestBlue}{HTML}{A7D8F0}
\definecolor{PearlSecondBlue}{HTML}{DCEFF8}
\definecolor{PearlThirdBlue}{HTML}{F2F8FC}
\newcommand{\bestcell}[1]{\cellcolor{PearlBestBlue}\textbf{#1}}
\newcommand{\secondcell}[1]{\cellcolor{PearlSecondBlue}#1}
\newcommand{\thirdcell}[1]{\cellcolor{PearlThirdBlue}#1}
\newcolumntype{L}{>{\raggedright\arraybackslash}p{0.26\linewidth}}
\newcolumntype{Y}{>{\raggedright\arraybackslash}X}
\newcommand{\sg}{\operatorname{sg}}

\title{PearlVLA: Progressive Embodied Action-Plan Refinement in Latent Space}

\author{
Bochen Yang\thanks{Corresponding author: \texttt{b.yang22@alumni.imperial.ac.uk}}\\
Imperial College London
\And
Lianlei Shan\\
Tsinghua University
}

\begin{document}

\maketitle

\begin{abstract}
Current Vision-Language-Action (VLA) models face a trade-off between efficient action generation and explicit deliberation.
Directly decoding actions from vision-language backbone representations enables low-latency control,
whereas textual reasoning, pixel-level subgoals, or world-model evaluation of decoded actions can improve planning but incur substantial latency and computational cost. 
We propose PearlVLA, a VLA framework that progressively refines a VLM-derived latent plan using feedback from the predicted consequence of each intermediate plan.
PearlVLA uses a frozen latent world model (LaWM) pretrained on action-free video.
At each refinement round, the current plan produces a continuous latent action code, and the LaWM predicts the corresponding latent visual subgoal.
A future-guided plan refiner uses this subgoal to update the plan, so each revision reshapes the next LaWM query.
After $K$ rounds, the refined plan is passed once to the host policy's action interface to produce an action chunk.
We further introduce Causal Refinement-Grouped Process-Reward RL to optimize latent refinement by comparing rewards from the longer-horizon imagined futures of plan edits made at the same refinement state.
Experiments on the LIBERO and RoboCasa benchmarks show that PearlVLA performs competitively against strong existing methods.
\end{abstract}

\section{Introduction}
\label{sec:introduction}

Vision-Language-Action models (VLAs) provide a unified framework for embodied
intelligence by mapping visual observations, language instructions, and robot
states to actions~\citep{rt2,pi_0}. Embodied manipulation, however, cannot be
reduced to isolated observation-to-action mappings. Over long horizons,
reasoning about scene geometry, object interactions, and future state
transitions is essential, as errors in grasp pose, motion direction, or goal
interpretation can compound into failure. A capable policy should therefore
anticipate the likely consequence of its current plan before
acting~\citep{unipi}.

\begin{figure}[t]
  \centering
  \includegraphics[width=\linewidth]{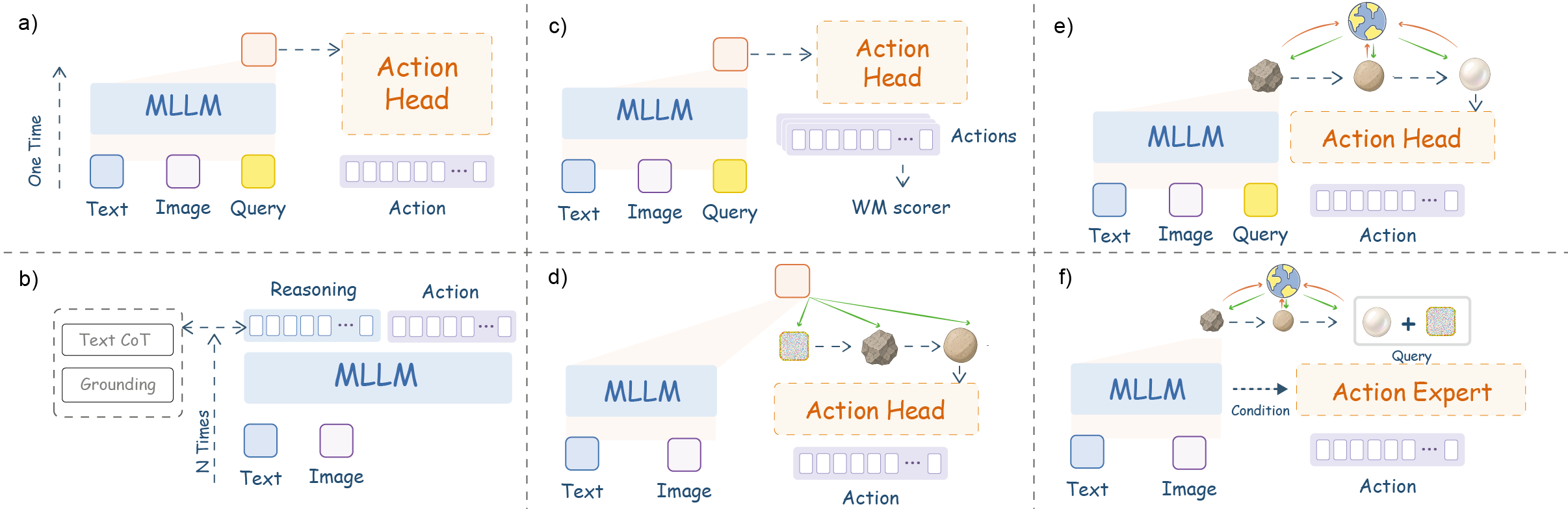}
  \caption{\textbf{Design choices for VLA policies.}
  (a) Direct action decoding.
  (b) Explicit intermediate reasoning.
  (c) Action-space world-model search.
  (d) Latent recurrent reasoning.
  (e) PearlVLA with a regression policy.
  (f) PearlVLA with a diffusion or flow policy.}
  \label{fig:intro}
\end{figure}

A natural way to provide foresight is to place an intermediate signal before
action decoding. Prior VLA systems have used textual reasoning traces or visual
subgoal generation~\citep{ecot,cot-vla}, feedback from value functions or world
models on decoded action proposals~\citep{hume,ladiwm}, and recurrent latent
computation~\citep{rdvla}. Although helpful, these mechanisms have different
limitations: explicit generation adds inference before action decoding,
action-level feedback ties each candidate or revision to an action decode, and
latent recurrence receives no modeled consequence between updates. Efficient
VLA methods avoid these intermediate stages by directly decoding actions from
VLM representations~\citep{oft}, but this direct path provides no explicit
mechanism to preview the future induced by the current plan or self-correct
before action decoding. Figure~\ref{fig:intro}(a--d) contrasts these design
choices with ours (e--f).

For embodied control, latent deliberation concerns not only internal computation but also the anticipated physical consequences of the plan being refined. Recent work learns
continuous latent action codes from visual transitions and uses a latent world
model (LaWM) to predict latent visual subgoals without pixel
reconstruction~\citep{garrido2026}. In existing single-query policies, the VLM
estimates the code without access to the future observation, and the resulting
subgoal conditions action generation rather than plan
refinement~\citep{lawam}.

To turn this one-way interface into a refinement loop, PearlVLA keeps the
current visual feature $u_t$ fixed while maintaining a writable VLM-derived
latent plan. At refinement round $k$, the committed latent action code $c_k$
queries the frozen LaWM for a latent visual subgoal $\hat{u}_k$. A
future-guided plan refiner uses this subgoal and its change from the first
prediction to update $z_{k-1}$ into $z_k$, which then composes the next query
$c_{k+1}$. Thus each revision reshapes the next LaWM query rather than merely
consuming a static future feature. After $K$ rounds, the refined plan is
consumed by the host policy's action interface to produce an action chunk. We
further introduce Causal Refinement-Grouped
Process-Reward RL (CRG-PRL), which treats each latent-plan edit as an action in
an inner decision process. For each candidate edit, the resulting latent visual
subgoal is autoregressively extended, rendered, and scored to obtain a process
reward. CRG-PRL compares edits sampled from the same refinement state,
separating edit quality from state difficulty through group-relative advantages
without a learned value function~\citep{shao2024deepseekmath,zheng2025group}.

Our contributions are summarized as follows:
\begin{enumerate}
    \item We present PearlVLA, a VLA framework for progressive, future-guided
    refinement of a VLM-derived latent action plan before action decoding.

    \item We introduce an anchored plan--code--future loop in which each plan
    update composes the next latent action code, elicits an updated modeled
    future from the frozen LaWM, and feeds that future into the next plan
    revision.

    \item Building on supervised refinement, we introduce CRG-PRL, a
    critic-free method that optimizes intermediate plan edits by comparing the
    longer-horizon imagined outcomes of alternative edits from the same
    refinement state.

    \item Experiments on LIBERO and RoboCasa show that PearlVLA performs
    competitively against strong baselines, achieving average success rates of
    98.7\% and 67.8\%, respectively.
\end{enumerate}

\section{Related Work}
\label{sec:related-work}

\paragraph{Reasoning and refinement in VLAs.}
Language-model research shows that useful intermediate computation need not be
fully verbalized. Coconut recurrently feeds hidden states back as continuous
thoughts, while looped Transformers obtain reasoning depth by repeatedly
applying shared computation~\citep{coconut,reasoningwithlatentthoughts}.
In VLAs, ECoT and CoT-VLA generate textual or visual
intermediates before actions~\citep{ecot,cot-vla}. Diffusion Policy and RD-VLA
spend additional inference computation on iterative action sampling or recurrent
latent states~\citep{diffusionpolicy,rdvla}. World-model feedback has also been incorporated into action refinement by LaDi-WM and ProgressVLA, with related work applying a single proposal-conditioned consequence update~\citep{ladiwm,progressvla,omegaeva}.

\paragraph{World models and latent actions.}
Predictive representations and latent actions provide compact dynamics
abstractions for robot policies. WoG distills future-conditioned features into
the VLM, while LAPO and UniVLA learn latent actions from visual transitions for
pretraining and control~\citep{wog,schmidt2023learning,bu2025univla}. LaWAM
retains the latent-action decoder as LaWM and conditions an action expert on its
predicted latent visual subgoal~\citep{lawam}. PearlVLA closes the latent
predictive loop at the plan level: each LaWM response revises the writable VLM
plan and thereby reshapes the next query, while the downstream action interface
is invoked only after refinement.

\section{Method}
\label{sec:method}

\begin{figure}[t]
  \centering
  \includegraphics[width=\linewidth]{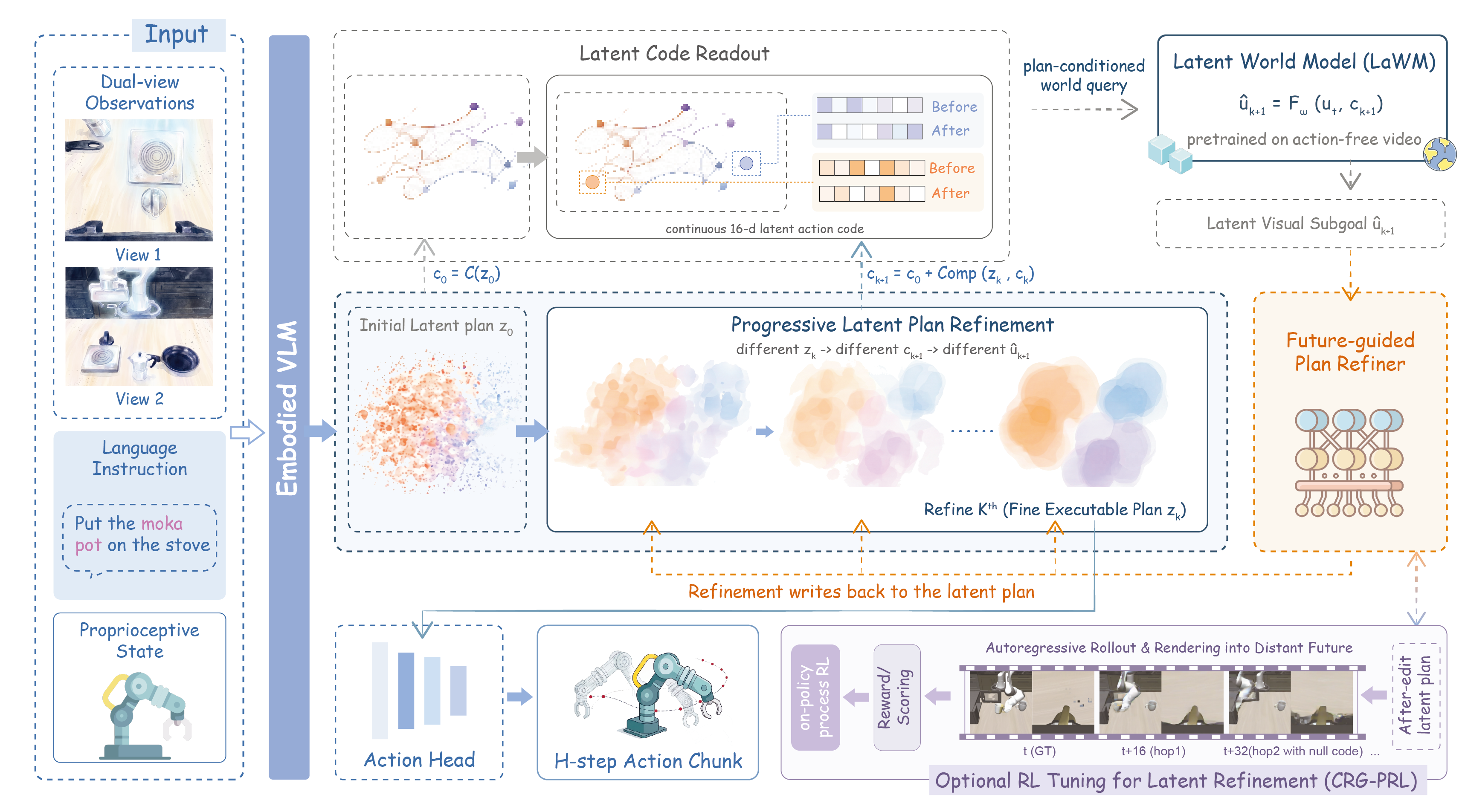}
  \caption{\textbf{The framework of PearlVLA.}
  The embodied VLM encodes its inputs into an initial latent plan.
  Over $K$ rounds, each refinement step composes a latent-code query, retrieves
  a latent visual subgoal from the frozen LaWM, and writes the resulting
  future feedback back into the plan. The refined plan is then decoded into an
  $H$-step action chunk. The optional CRG-PRL branch further optimizes the
  refinement process during training.}
  \label{fig:framework}
\end{figure}

Figure~\ref{fig:framework} provides an overview of PearlVLA. Given the current
RGB observation $o_t$, language instruction $\ell$, and proprioceptive state
$s_t$, a host-specific interface obtains an initial latent plan $z_0$ from the
embodied VLM context. PearlVLA then couples three representation spaces: the
writable VLM plan $z$, a continuous latent action code $c$, and the
visual-feature space $u$ of a frozen latent world model. At each round, the
current code queries LaWM for a corresponding latent future, and the resulting
feedback revises the plan before the next code is composed. After $K$ rounds, the host
action interface is conditioned on the refined plan together with its
instantiation-specific context. We reserve $z$ for the latent plan and $c$ for
the latent action code throughout.

For the OpenVLA instantiation, PearlVLA uses an OpenVLA-7B backbone following
the OpenVLA-OFT design~\citep{openvla,oft}. A transformer-based proprio encoder
converts proprioceptive history into tokens, which are fused with visual and
language tokens in the multimodal sequence
\texttt{[BOS] [vision] [proprio] [text] [meta\_query]}; the final
representations of eight meta-query tokens provide the OpenVLA plan
initialization. For GR00T
N1.5~\citep{bjorck2025gr00t}, sixteen learnable plan queries cross-attend to
the frozen VLM's instruction-token states following the three image-token
blocks, and the resulting query representations provide the GR00T plan
initialization. The
subsequent plan--code--prediction loop is shared across both hosts.

\subsection{Latent World Model as a Predictive Interface}
\label{sec:method-lam}

PearlVLA obtains its predictive interface from a latent action model pretrained
on videos without robot-action inputs~\citep{garrido2026}. Let
$o_t^{\mathrm{wm}}\subseteq o_t$ denote the two-view subset shared by LaWM,
the renderer, and the reward branch; benchmark-specific view choices are
given in Section~\ref{sec:experimental-setup}. A frozen visual encoder
$E_\psi$ maps the current and horizon view pairs to $u_t$ and
$u_{t+\Delta}$. An inverse-dynamics encoder $q_\varphi$ compresses their visual
transition into a continuous 16-dimensional latent action code, while a
forward-dynamics decoder $F_\omega$ predicts the corresponding horizon
feature:
\begin{equation}
\begin{aligned}
u_t&=E_\psi(o_t^{\mathrm{wm}}),&
u_{t+\Delta}&=E_\psi(o_{t+\Delta}^{\mathrm{wm}}),\\
c&\sim q_\varphi(c\mid u_t,u_{t+\Delta}),&
\hat u_{t+\Delta}&=F_\omega(u_t,c).
\end{aligned}
\label{eq:method-lam}
\end{equation}
Here $\Delta$ denotes the visual prediction offset of the latent action model;
our experiments use $\Delta=16$ frames.
After pretraining, $q_\varphi$ and $F_\omega$ are frozen, and $F_\omega$ serves
as the latent world model (LaWM). LaWM is conditioned on the transition code
$c$ and returns a latent visual subgoal $\hat u_{t+\Delta}$. During policy
training, the posterior mean
$c^\star=\mathbb E_{q_\varphi(c\mid u_t,u_{t+\Delta})}[c]$ provides a hindsight
target. At test time, the future feature is unavailable, so a host readout
infers the VLM-derived initial foresight code $c_0$ from current context alone.
Figure~\ref{fig:latent-code} illustrates
the sensitivity of the frozen predictive interface to teacher, initial, and
intermediate codes; refinement ablations evaluate the utility of repeated
queries. LaWM pretraining and the training-only renderer are detailed in
Appendix~\ref{app:lawm}.

\subsection{Anchored Future-Guided Plan Refinement}
\label{sec:method-refinement}

The central role of the predicted future is to provide an update signal for
the latent plan. Let $C$ denote the host-specific readout from the initial plan
to the LaWM code space. For the initial readout only, let $\tilde z_0$ denote
the VLM plan underlying the writable initial state $z_0$;
host-specific initialization details are deferred to
Appendix~\ref{app:policy}. The VLM-derived initial foresight code
$c_0=C(\tilde z_0)$ receives direct code supervision, while its uses inside
the refinement loop are detached. Its predicted subgoal likewise provides a
detached reference:
\begin{equation}
c_0=C(\tilde z_0),\qquad
\hat u_0=\sg\!\left(F_\omega(u_t,c_0)\right),\qquad
c_1=\sg(c_0)+\operatorname{Composer}(z_0,\sg(c_0)).
\label{eq:refinement-init}
\end{equation}
For $k=1,\ldots,K$, we write
$\hat u_k\equiv\hat u_{t+\Delta}^{(k)}=F_\omega(u_t,c_k)$, where $k$ indexes
the refinement query rather than physical time; every query predicts the same
offset $\Delta$.
At refinement round $k=1,\ldots,K$, the committed code $c_k$ queries LaWM for
a latent visual subgoal. FutureEncoder summarizes the modeled transition from
the current observation together with its change from the detached reference
subgoal, and RefineNet maps this relative future feedback into a residual plan
edit:
\begin{equation}
\begin{aligned}
\hat u_k
&=F_\omega(u_t,c_k),\\
\tau_k
&=\left[\hat u_k-u_t\,;\,\hat u_k-\hat u_0\right],
\qquad
h_k=\operatorname{FutureEncoder}(\tau_k),\\
\delta_k
&=\operatorname{RefineNet}(z_{k-1},h_k,c_k),\\
z_k
&=z_{k-1}+w_k\delta_k,\\
c_{k+1}
&=\sg(c_0)+\operatorname{Composer}(z_k,c_k).
\end{aligned}
\label{eq:refinement-loop}
\end{equation}
The first component of $\tau_k$ represents the visual change predicted under
the current code, while the second measures its deviation from the shared
reference subgoal $\hat u_0$. Composer uses the revised plan and current code
to form the next query relative to $\sg(c_0)$, avoiding an explicit cumulative
update from $c_k$ while retaining the code history. Fixed round-wise
coefficients $w_k$ control the residual write-back. Although LaWM remains
frozen, gradients through each round-dependent prediction train the modules
that generate subsequent queries. After $K$ rounds, $z_K$ combines the initial
VLM context with a sequence of future-guided revisions and serves as a central
conditioning signal for host action generation.

\subsection{Host-Policy Action Learning}
\label{sec:method-action}

After $K$ rounds, the host action interface receives the refined plan:
\begin{equation}
\hat y_{\mathrm{host}}
=\mathcal D_{\mathrm{host}}(z_K;\chi_t),
\label{eq:host-action}
\end{equation}
where $\chi_t$ collects host-specific context other than the latent plan.
Here $\hat y_{\mathrm{host}}$ denotes the host's supervised prediction: an
action chunk for OpenVLA and a flow-matching velocity for GR00T N1.5. At
deployment, the former is regressed in parallel, whereas the latter is
integrated by the host-native flow procedure. In both cases, the host interface
combines $z_K$ with its native context $\chi_t$ to produce an $H$-step action
chunk.

Supervised training combines the host objective with compact code-space
supervision:
\begin{equation}
\mathcal L_{\mathrm{SFT}}
=
\lambda_{\mathrm{host}}\,
\ell_{\mathrm{host}}(\hat y_{\mathrm{host}},y_{\mathrm{host}}^\star)
+\lambda_{\mathrm{init}}\operatorname{MSE}(c_0,c^\star)
+\lambda_{\mathrm{ref}}\operatorname{MSE}(c_{K+1},c^\star).
\label{eq:sft-objective}
\end{equation}
The host term trains the refined plan for downstream control, while the
initial and terminal code terms calibrate the LaWM interface before and after
refinement. Implementation details, loss weights, and benchmark-specific
settings are given in Appendix~\ref{app:policy}.

\subsection{Causal Refinement-Grouped Process-Reward RL}
\label{sec:method-crg-prl}

The supervised objective constrains the final action and endpoint codes, but
does not directly rank alternative edits at an intermediate refinement state.
We therefore introduce Causal Refinement-Grouped Process-Reward RL (CRG-PRL).
Rather than optimizing a multi-round return, CRG-PRL treats each refinement
round as a round-indexed local decision with $\gamma=0$. Define the inner state
$x_k=(z_{k-1},h_k,c_k,k)$. RefineNet receives the first three components; the
round index selects the prescribed write-back and exploration scales $w_k$ and
$\sigma_k$. We write its trainable bottleneck mean as $\mu_\theta(x_k)$ and
freeze the following up-projection $U_{\mathrm{up}}$. The resulting Gaussian
bottleneck policy and post-edit transition are
\begin{equation}
\begin{aligned}
b_k\sim\pi_\theta^b(\cdot\mid x_k)
&=\mathcal N\!\left(\mu_\theta(x_k),\sigma_k^2I\right),&
\delta_k&=U_{\mathrm{up}}b_k,\\
z_k&=z_{k-1}+w_k\delta_k,&
c_{k+1}&=\sg(c_0)+\operatorname{Composer}(z_k,c_k).
\end{aligned}
\label{eq:crg-policy-main}
\end{equation}

\paragraph{Post-edit process reward.}
The pre-edit prediction $\hat u_k=F_\omega(u_t,c_k)$ is formed before $b_k$ is
sampled and is therefore action-independent within a same-state group.
CRG-PRL instead scores the consequence induced by the post-edit code
$c_{k+1}$. Let
\begin{equation}
\hat u_{k,1}^{\mathrm{rew}}
=F_\omega(u_t,c_{k+1}),\qquad
\hat u_{k,2}^{\mathrm{rew}}
=F_\omega(\hat u_{k,1}^{\mathrm{rew}},c_{\mathrm{null}}).
\label{eq:crg-reward-futures-main}
\end{equation}
Here $c_{\mathrm{null}}$ is the learned no-code slot.
The frozen renderer converts these two supported hops into the ordered
dual-view sequence
$O_k=[o_t^{\mathrm{wm}},
\hat o_{k,1}^{\mathrm{rew}},\hat o_{k,2}^{\mathrm{rew}}]$.
With $\zeta_t$ collecting the fixed observation, instruction, and reward-side
context, the causal assignment is
\begin{equation}
\begin{gathered}
b_k\rightarrow\delta_k\rightarrow z_k\rightarrow c_{k+1}
\rightarrow
(\hat u_{k,1}^{\mathrm{rew}},\hat u_{k,2}^{\mathrm{rew}})
\rightarrow O_k\rightarrow r_k,\\
r_k=\mathcal R_{\mathrm{proc}}(\zeta_t,c_{k+1}).
\end{gathered}
\label{eq:crg-post-edit-chain-main}
\end{equation}
For $k=K$, the same construction uses $c_{K+1}$ only in the training-time
reward branch; the host action remains decoded from $z_K$. The complete
renderer and Robometer definitions are given in
Appendix~\ref{sec:crg-reward}.

\paragraph{Same-state local branching.}
Each collection iteration first runs a deterministic base path under a rollout
snapshot $\theta_{\mathrm{roll}}$, producing shared states
$\{x_k^{\mathrm{base}}\}_{k=1}^K$. At every base state, we sample $M$ one-round
branches:
\begin{equation}
\begin{aligned}
b_k^{(i)}
&=\mu_{\theta_{\mathrm{roll}}}(x_k^{\mathrm{base}})
+\sigma_k\eta_k^{(i)},
\qquad
\eta_k^{(i)}\sim\mathcal N(0,I),\\
z_k^{(i)}
&=z_{k-1}^{\mathrm{base}}+w_kU_{\mathrm{up}}b_k^{(i)},\\
c_{k+1}^{(i)}
&=\sg(c_0)+\operatorname{Composer}
\!\left(z_k^{(i)},c_k^{\mathrm{base}}\right),\\
r_k^{(i)}
&=\mathcal R_{\mathrm{proc}}
\!\left(\zeta_t,c_{k+1}^{(i)}\right),
\qquad i=1,\ldots,M .
\end{aligned}
\label{eq:crg-branches-main}
\end{equation}
A branch is evaluated for one round and is not propagated into the next base
state. All candidates in the group therefore share the pre-edit state,
refinement round, and reward-side randomness; only the sampled bottleneck edit
varies. Their rewards consequently provide a controlled within-model
comparison of local edit quality.

\paragraph{Group-relative policy update.}
For each same-state group, let $\bar r_k$ and $s_k^r$ denote the population
mean and standard deviation of its $M$ rewards. The group-relative advantage
and compact policy objective are
\begin{equation}
\begin{aligned}
\bar r_k&=\frac{1}{M}\sum_{i=1}^M r_k^{(i)},\qquad
s_k^r=
\sqrt{\frac{1}{M}\sum_{i=1}^M
\left(r_k^{(i)}-\bar r_k\right)^2},\\
A_k^{(i)}
&=\sg\!\left[
\frac{r_k^{(i)}-\bar r_k}{s_k^r+\varepsilon}
\right],\qquad
\mathcal L_{\mathrm{PG}}
=-\mathbb E_{k,i}\!\left[
A_k^{(i)}
\log\pi_\theta^b\!\left(
\sg(b_k^{(i)})\mid\sg(x_k^{\mathrm{base}})
\right)\right].
\end{aligned}
\label{eq:crg-pg-main}
\end{equation}
Here $\varepsilon>0$ stabilizes groups with near-zero reward spread. The
same-state mean provides a local baseline without a learned critic,
following group-relative credit
assignment~\citep{shao2024deepseekmath,zheng2025group}. The complete objective
adds an SFT edit-proximity term and a supervised host-action anchor:
\begin{equation}
\mathcal L_{\mathrm{CRG}}
=\mathcal L_{\mathrm{PG}}
+\lambda_{\mathrm{prox}}\mathcal L_{\mathrm{prox}}
+\lambda_{\mathrm{act}}\mathcal L_{\mathrm{act}}.
\label{eq:crg-total-main}
\end{equation}
CRG-PRL uses fresh on-policy buffers and updates only the pre-expansion layers
of RefineNet. Reward-side branching is training-only, and deployment follows
the deterministic mean path. Here ``causal'' denotes a controlled
model-internal intervention that holds the pre-edit state and frozen proxy
pipeline fixed while varying only the bottleneck edit; it does not assert
physical causality in the real environment. Full estimator, numerical,
freezing, calibration, and guardrail details are provided in
Appendix~\ref{app:crg-prl}.

\FloatBarrier
\section{Experiments}
\label{sec:experiments}

\subsection{Experimental Setup}
\label{sec:experimental-setup}

\paragraph{LIBERO and LIBERO-Plus.}
For the OpenVLA-based instantiation, we evaluate PearlVLA on the standard
LIBERO benchmark~\citep{liu2023libero}, which comprises four generalization
suites: Spatial for spatial relations, Object for object-centric
generalization, Goal for goal-conditioned execution, and Long for long-horizon
task composition. We train on the RLDS-formatted modified LIBERO
dataset~\citep{openvla}, which repackages the official human-teleoperated
demonstrations from the four suites. Each suite contains 10 tasks and 50 expert
demonstrations per task (500 per suite), and provides camera images, robot
states, task instructions, and continuous delta end-effector actions. Unless
otherwise stated, we follow the standard suite-specific protocol: one policy
is trained separately for each suite, and the four suite success rates are
averaged after evaluation. Task success rate is the primary metric. We further
evaluate robustness on LIBERO-Plus~\citep{fei2025libero}, which extends the
four suites with 10,030 perturbed evaluation episodes spanning controlled
variations in layout, camera viewpoint, robot state, language, lighting,
background, and sensor noise.

\paragraph{RoboCasa.}
For the GR00T-based instantiation, we evaluate PearlVLA on the standard
24-task RoboCasa kitchen benchmark~\citep{nasiriany2024robocasa} under the
300-demonstration-per-task protocol, training one multi-task policy over all
24 tasks and evaluating 50 rollouts per task. We report task success rates for
Pick-and-Place, Open-or-Close, Others, and all tasks.

\paragraph{Policy instantiations.}
The LIBERO experiments use an OpenVLA-7B policy following the OpenVLA-OFT
design~\citep{openvla,oft}, with parallel eight-step action regression, whereas
the RoboCasa experiments use GR00T N1.5~\citep{bjorck2025gr00t} with its native
16-step flow-matching action expert; both use $K=3$ refinement rounds. During
evaluation, each policy executes its complete predicted action chunk
before the next policy query. For both instantiations, the
LaWM, renderer, and reward branch share two views:
third-person plus wrist for LIBERO, and external-left plus wrist for RoboCasa;
the RoboCasa host VLM additionally consumes a second third-person view. We use
frozen LaWM checkpoints pretrained on an equal
mixture of action-free robot videos from LIBERO, LIBERO-90, and RoboCasa and
then specialized to each benchmark, without using robot actions as model
inputs; the renderer and Robometer~\citep{liang2026robometer} are used only for
CRG-PRL reward construction and not at policy inference. Complete LaWM,
policy-training, and implementation details are provided in
Appendix~\ref{app:lawm} and
Tables~\ref{tab:libero-hparams}--\ref{tab:robocasa-hparams}.

\subsection{Results}
\label{sec:benchmark-results}

\paragraph{LIBERO.}
PearlVLA uses the same OpenVLA-7B initialization and parallel continuous
action-chunk output family as OpenVLA-OFT, but inserts $K=3$ future-guided
latent refinement rounds before its action head. The main table reports both
the supervised PearlVLA model and the final model after CRG-PRL tuning.
Although flow-matching action heads have been reported to outperform
regression heads on standard VLA benchmarks at the cost of iterative
inference~\citep{vlanext,pi_05}, PearlVLA keeps a parallel regression head and
instead allocates additional inference compute to plan refinement before
action decoding. We use OpenVLA-OFT~\citep{oft} as the primary external
reference because it is a strong LIBERO model with the same backbone
initialization and action-output family. Table~\ref{tab:libero} also includes
representative VLA baselines as broader reference points; these methods differ
in backbone, data mixture, action representation, and training recipe, so they
should not be read as fully controlled comparisons.

\begin{table}[!htbp]
  \caption{LIBERO suite-specific success rates (\%). Blue shades mark the
  three highest distinct score levels within the
  regression/classification VLA group; the final row reports PearlVLA after
  CRG-PRL tuning.}
  \label{tab:libero}
  \centering
  \normalsize
  \setlength{\tabcolsep}{5pt}
  \begin{tabular}{@{}lccccc@{}}
    \toprule
    Model & Spatial & Object & Goal & Long & Avg \\
    \midrule
    \rowcolor{PearlGroupGray}\multicolumn{6}{@{}l}{\emph{Flow-matching or diffusion-based VLA}} \\
    FLOWER~\citep{reuss2025flower} & 97.5 & 99.1 & 96.1 & 94.9 & 96.9 \\
    VLANeXt~\citep{vlanext} & 99.0 & 99.2 & 96.6 & 94.6 & 97.4 \\
    $\pi_{0.5}$~\citep{pi_05} & 97.0 & 99.0 & 98.0 & 96.0 & 97.5 \\
    ProgressVLA~\citep{progressvla} & 88.2 & 96.4 & 87.2 & 66.2 & 84.5 \\
    LaWAM~\citep{lawam} & 99.4 & 99.6 & 98.4 & 97.0 & 98.6 \\
    \midrule
    \rowcolor{PearlGroupGray}\multicolumn{6}{@{}l}{\emph{Regression or classification-based VLA}} \\
    OpenVLA~\citep{openvla} & 84.7 & 88.4 & 79.2 & 53.7 & 76.5 \\
    WorldVLA~\citep{cen2025worldvla} & 85.6 & 89.0 & 82.6 & 59.0 & 79.1 \\
    CoT-VLA~\citep{cot-vla} & 87.5 & 91.6 & 87.6 & 69.0 & 83.9 \\
    NORA~\citep{hung2025nora} & 92.2 & 95.4 & 89.4 & 74.6 & 87.9 \\
    RD-VLA~\citep{rdvla} & 92.0 & \secondcell{99.0} & 96.0 & 84.8 & 93.0 \\
    UniVLA~\citep{bu2025univla} & 96.5 & 96.8 & 95.6 & 92.0 & 95.2 \\
    $\pi_0$-Fast~\citep{pertsch2025fast} & 96.4 & 96.8 & 88.6 & 60.2 & 85.5 \\
    $\pi_0$(reg)~\citep{pi_0} & \thirdcell{97.8} & 98.2 & 94.6 & 90.2 & 95.2 \\
    OpenVLA-OFT~\citep{oft} & 97.6 & \thirdcell{98.4} & \thirdcell{97.9} & \thirdcell{94.5} & \thirdcell{97.1} \\
    \textbf{PearlVLA (Ours)} & \secondcell{99.2} & \bestcell{99.8} & \secondcell{98.2} & \secondcell{96.6} & \secondcell{98.5} \\
    \textbf{PearlVLA + CRG-PRL (Ours)} & \bestcell{99.4} & \bestcell{99.8} & \bestcell{98.4} & \bestcell{97.2} & \bestcell{98.7} \\
    \bottomrule
  \end{tabular}
\end{table}

Supervised PearlVLA improves over OpenVLA-OFT on all four suites, raising the
average success rate from 97.1\% to 98.5\%. CRG-PRL further raises the average
to 98.7\%, with the largest absolute gain over OpenVLA-OFT on LIBERO-Long
(2.7 points). Within the regression/classification group, the final model
obtains the highest average and suite-specific success rates. Among all
methods shown, it also edges the strongest flow-matching baseline, LaWAM,
by 0.1 points on average. Since the broader baselines use different backbones
and training recipes, these comparisons establish empirical rankings under
the reported protocols rather than controlled attribution of every gain.

\paragraph{Controlled robustness on LIBERO-Plus.}
To assess robustness beyond the in-domain setting, we additionally evaluate
supervised PearlVLA on LIBERO-Plus~\citep{fei2025libero}, which extends the
four LIBERO suites with 10,030 perturbed evaluation episodes spanning
controlled variations in layout, camera viewpoint, robot state, language,
lighting, background, and sensor noise.

\begin{table}[!htbp]
  \caption{LIBERO-Plus robustness under controlled perturbation settings.
  Results are shown in success rate (\%). PearlVLA results are averaged over
  perturbation suites and three inference seeds.}
  \label{tab:libero-plus}
  \centering
  \small
  \setlength{\tabcolsep}{3.5pt}
  \begin{tabular}{@{}lcccccccc@{}}
    \toprule
    Model & Camera & Robot & Language & Light & Background & Noise & Layout & Total \\
    \midrule
    OpenVLA~\citep{openvla} & 0.8 & 3.5 & 23.0 & 8.1 & 34.8 & 15.2 & 28.5 & 15.6 \\
    $\pi_0$-Fast~\citep{pertsch2025fast} & 65.1 & 21.6 & 61.0 & 73.2 & 73.2 & 74.4 & 68.8 & 61.6 \\
    OpenVLA-OFT~\citep{oft} & 56.4 & 31.9 & 79.5 & 88.7 & 93.3 & 75.8 & 74.2 & 69.6 \\
    \rowcolor{gray!15} PearlVLA (Ours) & 70.4 & 48.3 & 78.9 & 94.4 & 92.9 & 79.8 & 79.3 & 76.3 \\
    \bottomrule
  \end{tabular}
\end{table}

PearlVLA improves over OpenVLA-OFT from 69.6\% to 76.3\% total success. The
largest gains occur under camera and robot-state perturbations. These results complement standard LIBERO
and suggest that latent refinement improves generalization under controlled perturbations
beyond the in-domain protocol.

\paragraph{RoboCasa.}
To test whether the refinement mechanism is tied to the regression head and
meta-query interface used on LIBERO, we instantiate PearlVLA on GR00T
N1.5~\citep{bjorck2025gr00t}, where a frozen VLM conditions a flow-matching
action expert. The host-specific plan readout and clean-prefix injection adapt
the interface while retaining the same plan--code--prediction loop
(Appendix~\ref{app:policy}).

Our controlled host-policy baseline is vanilla GR00T N1.5 trained on the same
supervised RoboCasa data, but without PearlVLA's pretrained LaWM or refinement
modules. Two recent methods are included only as broader, non-controlled
references.

\begin{table}[!htbp]
  \caption{Results on the standard RoboCasa Kitchen 24-task benchmark with
  300 demonstrations per task. Cited rows are reported by prior work; GR00T
  N1.5 and PearlVLA variants are evaluated in our implementation.}
  \label{tab:robocasa}
  \centering
  \small
  \setlength{\tabcolsep}{6pt}
  \begin{tabular}{@{}lcccc@{}}
    \toprule
    Method & Pick-and-Place & Open-or-Close & Others & All \\
    \midrule
    Video Policy~\citep{liang2025video} & 47.2 & 83.0 & 70.0 & 65.7 \\
    GR00T N1.5 + HAMLET~\citep{koo2025hamlet} & 48.5 & 87.7 & 68.0 & 66.4 \\
    \midrule
    GR00T N1.5 (vanilla) & 51.0 & 82.3 & 66.8 & 65.4 \\
    PearlVLA ($K{=}1$) & 50.3 & 86.0 & 68.2 & 66.7 \\
    PearlVLA ($K{=}3$) & 52.5 & 88.0 & 67.2 & 67.5 \\
    PearlVLA ($K{=}3$) + CRG-PRL & 53.0 & 87.7 & 67.8 & \textbf{67.8} \\
    \bottomrule
  \end{tabular}
\end{table}

PearlVLA improves over vanilla GR00T N1.5 from 65.4\% to 66.7\% with one
refinement round and 67.5\% with the default three rounds. CRG-PRL further
raises the overall success rate to 67.8\%. The gains at both evaluated
refinement depths indicate that the same future-guided plan refinement remains
useful when the final plan conditions a flow-matching action expert rather
than a parallel regression head.

\FloatBarrier

\subsection{Ablations and Mechanistic Analysis}
\label{sec:ablations}

To separate fixed conditioning from plan-conditioned predictive feedback, we
evaluate controlled supervised OpenVLA regression policies on LIBERO. The
fixed-VLM control uses GR00T N1.5 output only as round-invariant RefineNet keys
and values; the one-shot LaWM control concatenates latent tokens from one
prediction, and the cached-LaWM control reuses one prediction across refinement
rounds. PearlVLA instead re-queries LaWM from the current plan-derived code.
Each value of $K$ denotes a separately trained fixed-depth model.

\begin{table}[!htbp]
  \caption{\textbf{Controlled mechanism ablation on LIBERO.}
  All variants use the OpenVLA regression policy under the same training and
  evaluation protocol, without CRG-PRL.
  $Q_{\mathrm{LaWM}}$ counts LaWM evaluations per policy query.
  The gray row marks the default supervised PearlVLA model.
  Results are four-suite average success rates (\%).}
  \label{tab:mechanism-ablation}
  \centering
  \small
  \setlength{\tabcolsep}{3.5pt}
  \begin{tabularx}{\linewidth}{@{}llYlccc@{}}
    \toprule
    Variant & Integration & Auxiliary signal & Signal update
      & $K$ & $Q_{\mathrm{LaWM}}$ & Avg. $\uparrow$ \\
    \midrule
    Direct control
      & Direct decoding
      & None
      & -- & 0 & 0 & 96.8 \\
    \midrule
    Fixed-VLM control
      & Refiner K/V
      & GR00T N1.5 VLM output
      & Fixed & 1 & 0 & 97.2 \\
    Fixed-VLM control
      & Refiner K/V
      & GR00T N1.5 VLM output
      & Fixed & 3 & 0 & 97.3 \\
    \midrule
    One-shot LaWM
      & Latent-token concat.
      & LaWM prediction
      & Once & 0 & 1 & 97.7 \\
    Cached-LaWM control
      & Refiner K/V
      & LaWM prediction
      & Once, then reused & 1 & 1 & 97.6 \\
    Cached-LaWM control
      & Refiner K/V
      & LaWM prediction
      & Once, then reused & 3 & 1 & 97.9 \\
    \midrule
    PearlVLA
      & Refiner K/V
      & LaWM prediction
      & From current plan & 1 & 2 & 97.8 \\
    \rowcolor{gray!15}
    PearlVLA
      & Refiner K/V
      & LaWM prediction
      & From current plan & 3 & 4 & \textbf{98.5} \\
    PearlVLA
      & Refiner K/V
      & LaWM prediction
      & From current plan & 5 & 6 & 98.4 \\
    \bottomrule
  \end{tabularx}
\end{table}

Direct decoding obtains 96.8\%. From $K=1$ to $K=3$, the fixed-VLM control
changes from 97.2\% to 97.3\%, and the cached-LaWM control from 97.6\% to
97.9\%, whereas plan-conditioned feedback improves from 97.8\% to 98.5\%.
Its advantage over cached feedback therefore grows from 0.2 to 0.6 points,
consistent with additional rounds being most useful when the evolving plan
changes the next predictive query. One-shot latent-token concatenation reaches
97.7\%, while the $K=5$ PearlVLA model reaches 98.4\%, suggesting saturation
around three rounds. Because plan-conditioned variants also use more LaWM
evaluations than the cached controls, this is an end-to-end mechanism
comparison rather than a compute-matched causal estimate.

\paragraph{Inference cost.}
The default $K=3$ loop increases policy-query latency from 156.6 to 198.7\,ms
on OpenVLA and from 59.0 to 101.0\,ms on GR00T N1.5, while adding 391 GFLOPs
in each case. Peak memory increases by 0.9 and 0.8\,GB, respectively, at the
first refinement round and remains unchanged through $K=3$; the renderer and
reward model are training-only. Full $K$-wise measurements are reported in
Tables~\ref{tab:eff-openvla} and~\ref{tab:eff-groot}.

\section{Conclusion}

We introduced PearlVLA, a VLA framework that turns a one-shot latent
world-action interface into a closed deliberation loop. Rather than repeatedly
processing a fixed condition or revising actions that have already been
decoded, PearlVLA refines a latent plan before action generation. Each revision
changes the consequence considered by the next round, while the host action
interface is invoked only after deliberation is complete. We further introduced
CRG-PRL, which provides process-level post-training for the intermediate
revisions themselves.

Across LIBERO, LIBERO-Plus, and RoboCasa, PearlVLA improves both regression-
and flow-based host policies and performs competitively with strong
contemporary methods. Controlled comparisons show the strongest gains when the
modeled consequence evolves with the plan, rather than remaining fixed or
being reused across rounds, while our efficiency analysis makes the
performance--cost trade-off explicit. Together, these results position
consequence-guided latent plan revision as a complementary direction to
recurrent latent computation, increasingly expressive action decoders, and
action-space search.

The present study uses a fixed refinement depth and benchmark-specialized LaWM
checkpoints trained on a limited robot-video mixture. Adaptive halting could
allocate deliberation on demand. More broadly, progress in large-scale
representation-space video prediction and latent action world modeling
suggests that pretraining on substantially broader action-free interaction
video could yield a more transferable predictive interface across tasks and
environments~\citep{assran2025v,garrido2026}. Validating this direction in
contact-rich real-world manipulation remains important.

\FloatBarrier
\newpage

\bibliographystyle{unsrtnat}
\bibliography{main}

\newpage
\appendix

\section{Latent World Model}
\label{app:lawm}

\subsection{LaWM}

PearlVLA uses the frozen forward decoder of a latent action model as its latent
world model (LaWM). A frozen DINOv3 ViT-B/16 visual encoder
$E_\psi$~\citep{simeoni2025dinov3} maps the two-view observations
$o_t^{\mathrm{wm}}$ and $o_{t+\Delta}^{\mathrm{wm}}$ into features $u_t$ and
$u_{t+\Delta}$. The inverse-dynamics encoder $q_\varphi$ and forward-dynamics
decoder $F_\omega$ are trained jointly:
\begin{equation}
c \sim q_\varphi(c\mid u_t,u_{t+\Delta}),
\qquad
\hat u_{t+\Delta}=F_\omega(u_t,c).
\label{eq:app-lam}
\end{equation}
The encoder compresses the observed visual transition into a continuous
16-dimensional latent action code, while the decoder predicts the corresponding
future features from the current features and that code. After pretraining,
both $q_\varphi$ and $F_\omega$ are frozen: $q_\varphi$ supplies the hindsight
teacher code $c^\star$ during policy training, while $F_\omega$ is queried as
LaWM; robot actions are not among its inputs. A separately trained renderer
$R_\xi$, written here as its complete inference map, predicts a future VAE
latent from the current-frame VAE latent and a LaWM visual subgoal:
\begin{equation}
v_t=E_{\mathrm{vae}}(o_t^{\mathrm{wm}}),\qquad
\hat v_{t+\Delta}=R_\xi(v_t,\hat u_{t+\Delta}),\qquad
\hat o_{t+\Delta}=D_{\mathrm{vae}}(\hat v_{t+\Delta}).
\label{eq:app-renderer}
\end{equation}
For a second hop, $\hat v_{t+\Delta}$ becomes the renderer's present input.
Rendering is used only for visualization and reward computation, not during
policy deployment.

\subsection{Latent-Code Geometry and Rendered Predictions}

\begin{figure}[!ht]
\centering
\includegraphics[width=\linewidth]{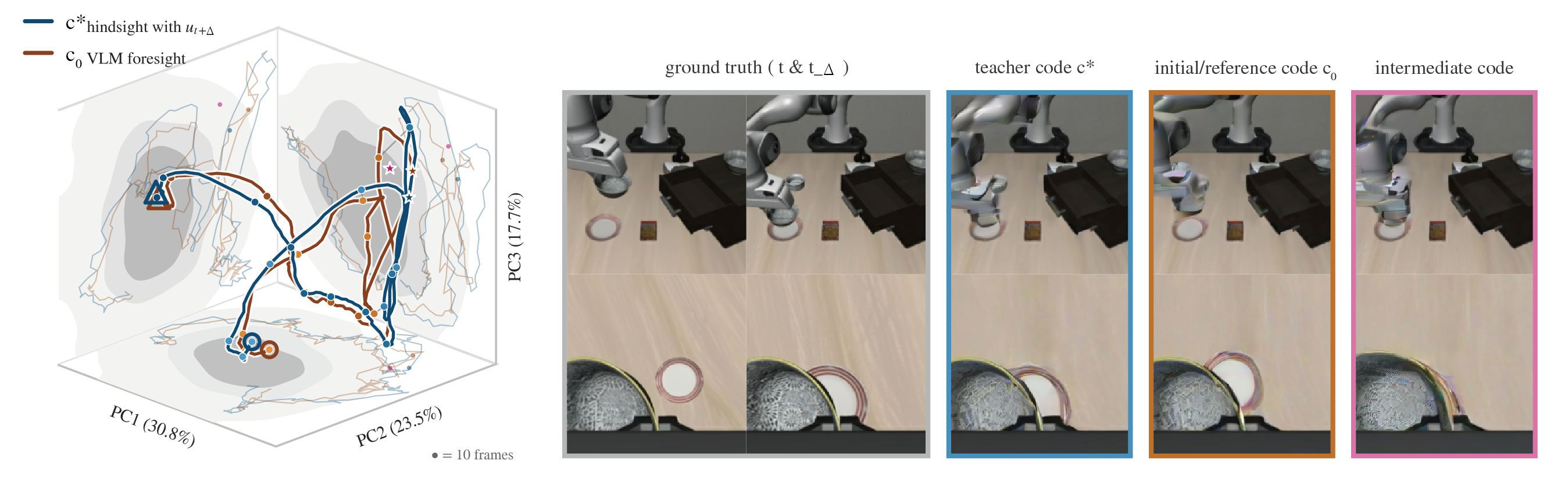}
\caption{\textbf{Geometry of the latent action code space and rendered visual
transitions.} \textit{Left:} Three-dimensional PCA trajectories of the
hindsight teacher code $c^\star$ and VLM foresight code
$c_0$. \textit{Right:} Ground-truth observations, followed by visual
subgoals predicted by the frozen LaWM $F_\omega$ under $c^\star$, $c_0$, and
an intermediate candidate code, then rendered by $R_\xi$.}
\label{fig:latent-code}
\end{figure}

Figure~\ref{fig:latent-code} provides two qualitative checks on the predictive
interface. The left panel projects trajectories of the VLM foresight code
$c_0$ and hindsight teacher code $c^\star$ into three principal components.
Their separation visualizes the partial-information gap: the VLM infers $c_0$
from the current observation and instruction, whereas $c^\star$ is encoded from
the observed transition $(u_t,u_{t+\Delta})$. The right panel queries the same frozen LaWM and
renderer with $c^\star$, $c_0$, and an intermediate candidate. The resulting
predictions are distinct yet locally coherent, which is consistent with
code sensitivity and local coherence in the frozen predictive interface. This
visualization motivates repeated
plan-conditioned queries; performance along the actual refinement sequence is
established by the refinement ablations rather than by the visualization alone.

\FloatBarrier

\subsection{Training Configuration}

LaWM pretraining first uses an equal mixture of LIBERO, LIBERO-90, and RoboCasa
videos, and then specializes separate checkpoints for LIBERO and RoboCasa. The
training objective combines horizon-feature prediction with a weak KL
regularizer. Both LaWM and the renderer are trained with a two-hop extension.
For LaWM, the first-hop prediction becomes the next current feature and
50\% of second-hop conditions use $c_{\mathrm{null}}$. For the renderer, the
first predicted VAE latent becomes the next ``present'' input, paired with the
second-hop subgoal predicted by LaWM with $c_{\mathrm{null}}$.
Table~\ref{tab:lawm-training} summarizes the configuration.

\begin{table}[!ht]
\caption{Training configuration of LaWM and the subgoal renderer.}
\label{tab:lawm-training}
\centering
\small
\begin{tabularx}{\linewidth}{@{}LY@{}}
\toprule
Setting & Value \\
\midrule
\rowcolor{PearlGroupGray}\multicolumn{2}{@{}l}{\textit{LaWM pretraining}} \\
Pretraining data & LIBERO, LIBERO-90, and RoboCasa videos; no robot actions as model inputs \\
Training curriculum &
30K mixed; then to 50K (LIBERO) / 60K (RoboCasa) \\
Visual input & Two $224\times224$ views; 16-frame offset \\
Backbone and code & Frozen DINOv3 ViT-B/16; continuous 16-D code \\
IDM / FDM & 12-layer encoder / 16-layer residual AdaLN decoder (width 768) \\
Objective & DINO-feature MSE $+\,10^{-5}$ KL \\
Two-hop extension &
First-hop prediction reused as the next current feature; 50\% use $c_{\mathrm{null}}$ \\
Optimization & AdamW; learning rate $2\times10^{-4}$; global batch size 72 \\
\midrule
\rowcolor{PearlGroupGray}\multicolumn{2}{@{}l}{\textit{Subgoal renderer}} \\
Model and objective &
Frozen SDXL VAE~\citep{podell2024sdxl}; 12-layer DiT; future VAE-latent noise
prediction \\
Two-hop extension &
First predicted VAE latent reused as the next present input; paired with a
$c_{\mathrm{null}}$-conditioned LaWM subgoal \\
Optimization &
AdamW; learning rate $2\times10^{-4}$; global batch size 64; 60K steps \\
\bottomrule
\end{tabularx}
\end{table}
\FloatBarrier

\section{Policy Architecture and Implementation Details}
\label{app:policy}

\paragraph{Shared refinement modules.}
Composer is implemented as cross-attention whose query combines a normalized
refinement state with a projection of the current code and attends to the
latent-plan tokens. FutureEncoder maps the visual tokens of $\tau_k$ to future
tokens, adding group, view, and patch embeddings. RefineNet projects the
normalized plan state to width 384 and adds a learned projection of $c_k$ to
each plan-token query, then cross-attends to the future tokens, which serve as
keys and values. A linear up-projection returns one 4096-dimensional residual
$\delta_k$ for each plan token. In practice, the write-back of
Eq.~\ref{eq:refinement-loop}, also used in CRG-PRL, normalizes this residual to
a fixed step size,
$z_k=z_{k-1}+w_k\,\sg(\operatorname{RMS}(z_0))\,\delta_k/\operatorname{RMS}(\delta_k)$,
with RMS taken per sample over all plan tokens and channels. Each round thus
changes the plan by $w_k\operatorname{RMS}(z_0)$ in RMS, with RefineNet
determining the direction of the edit.

\subsection{OpenVLA / LIBERO}

\begin{table}[!ht]
\setlength{\abovecaptionskip}{4pt}
\renewcommand{\arraystretch}{0.94}
\caption{Core supervised-policy settings for LIBERO.}
\label{tab:libero-hparams}
\centering
\small
\begin{tabularx}{\linewidth}{@{}LY@{}}
\toprule
Hyperparameter & Value \\
\midrule
\rowcolor{PearlGroupGray}\multicolumn{2}{@{}l}{\textit{Optimization}} \\
Optimizer & AdamW \\
Learning rate & $2\times10^{-4}$, cosine decay to $0.05\times$ the base rate \\
Global batch size & 32 \\
Training steps & 20K for Spatial/Object; 24K for Goal/Long \\
\midrule
\rowcolor{PearlGroupGray}\multicolumn{2}{@{}l}{\textit{Training objectives}} \\
Action regression & Normalized action-chunk MSE: $z_K$ head $\times25$, auxiliary $z_0$ head $\times5$ \\
Initial-code distillation & $c_0=C(\tilde z_0)$ vs.\ target $c^\star$, MSE scaled by 0.4 \\
Refined-code supervision & Terminal $c_{K+1}$ ($c_4$ for $K=3$) vs.\ target $c^\star$, MSE scaled by 0.1 \\
\midrule
\rowcolor{PearlGroupGray}\multicolumn{2}{@{}l}{\textit{Inputs and action output}} \\
Base VLA & OpenVLA-7B with LoRA rank 16 \\
Observation inputs &
Third-person and wrist images, language, and 8-step proprioception \\
Image resolution & $224\times224$ pixels per view \\
Image augmentation & 93.3\% random crop and color jitter \\
Action output & 7-dimensional actions in an 8-step chunk \\
Action decoder & Query-transformer regression head \\
\midrule
\rowcolor{PearlGroupGray}\multicolumn{2}{@{}l}{\textit{Latent refinement}} \\
Refinement rounds & $K=3$ \\
Plan perturbation & Gaussian, scaled by the plan RMS; $\rho_{\mathrm{init}}=0.1$ \\
Plan update & Fixed round-wise scalars with nominal update RMS of 15\%, 6\%, and 3\% \\
\midrule
\rowcolor{PearlGroupGray}\multicolumn{2}{@{}l}{\textit{Compute}} \\
Hardware & $4\times$ NVIDIA A100 40GB GPUs \\
\bottomrule
\end{tabularx}
\end{table}

PearlVLA adopts the OpenVLA-OFT architecture~\citep{openvla,oft}: an
OpenVLA-7B backbone with
fused SigLIP--DINOv2 features~\citep{oquab2023dinov2}, an LLaMA-2 7B language
model~\citep{touvron2023llama}, and a three-layer GELU MLP projector, adapted
via LoRA for parallel action-chunk regression.
A transformer-based proprio encoder converts proprioceptive history into
tokens, which are fused with the visual and language inputs in the multimodal
sequence \texttt{[BOS] [vision] [proprio] [text] [meta\_query]}. Eight learnable
meta-query tokens use a prefix-LM attention mask, and their final-layer
representations, with a small Gaussian perturbation scaled by their RMS
($\rho_{\mathrm{init}}=0.1$), form the initial plan $\tilde z_0$.

$c_0=C(\tilde z_0)$ and $\hat u_0=\sg(F_\omega(u_t,c_0))$ are computed before
refinement, and all paths from $c_0$ into the refinement loop use $\sg(c_0)$.
Applied after the trainable VLM, this perturbation serves as a mild
regularizer. Over $K$ rounds, the code queries the frozen LaWM for a DINOv3
latent subgoal, and the refiner applies scheduled residual updates from the
plan, the subgoal, and its change relative to the detached initial prediction
$\hat u_0$. An AdaLN-conditioned query transformer then regresses the
eight-step action chunk in parallel. An auxiliary head of the same form, used
only during training, regresses the chunk from the initial plan $z_0$. It
encourages the latent plan itself to be action-decodable, so that refinement
starts from an informative plan, while the higher-weighted action loss on the
refined plan trains the loop. Settings are reported in
Table~\ref{tab:libero-hparams}.

\FloatBarrier
\subsection{GR00T N1.5 / RoboCasa}

For RoboCasa, PearlVLA is built on GR00T N1.5~\citep{bjorck2025gr00t}. Its
Eagle-2.5 VLM, comprising a SigLIP vision encoder and a Qwen3 language model,
remains frozen; no LoRA is applied and no VLM weight is updated. A lightweight
query transformer uses sixteen learnable plan queries to cross-attend to the
frozen VLM's instruction-token states following the three image-token blocks,
producing sixteen latent-plan tokens. A Q-Former-style readout maps these
tokens to the same 16-dimensional LaWM code space.
The plan is refined through the same $K$-round LaWM loop used by the OpenVLA
instantiation.

Action decoding uses the native GR00T N1.5 flow-matching expert, initialized
from its pretrained weights and fully fine-tuned. The refined plan is projected
to the expert width and inserted as a separate clean prefix beside the
proprioceptive prefix. Noised action queries remain the variables of the
flow-matching objective, conditioned on the frozen VLM output through
cross-attention. The interpolation schedule, flow-time conditioning, velocity
target, and four-step Euler integrator are unchanged, and the expert outputs a
16-step, 12-dimensional mobile-manipulation action chunk.

During training, the complete VLM key-value sequence is replaced by a shared
learned null sequence with probability 0.10. This sample-level dropout keeps
the pretrained expert from ignoring the latent-plan prefix. Dropout is disabled
at inference.
Table~\ref{tab:robocasa-hparams} summarizes the training settings.

\begin{table}[!ht]
\caption{Core supervised-policy settings for RoboCasa.}
\label{tab:robocasa-hparams}
\centering
\small
\begin{tabularx}{\linewidth}{@{}LY@{}}
\toprule
Hyperparameter & Value \\
\midrule
\rowcolor{PearlGroupGray}\multicolumn{2}{@{}l}{\textit{Optimization}} \\
Optimizer & AdamW \\
Learning rate & $1\times10^{-4}$, followed by cosine decay to $0.05\times$ the base rate \\
Global batch size & 32 \\
Training steps & 60K \\
\midrule
\rowcolor{PearlGroupGray}\multicolumn{2}{@{}l}{\textit{Training objectives}} \\
Action regression & Flow-matching velocity MSE $\times10$; auxiliary $z_0$ head $\times3$ \\
Initial-code distillation &
MSE between $c_0$ and the 16-dim target code $c^\star$, scaled by 0.4 \\
Refined-code supervision &
MSE between terminal $c_{K+1}$ ($c_4$ for $K=3$) and $c^\star$, scaled by 0.1 \\
\midrule
\rowcolor{PearlGroupGray}\multicolumn{2}{@{}l}{\textit{Inputs and action output}} \\
Base VLA &
GR00T N1.5 with a frozen Eagle-2.5 VLM; no LoRA; fully fine-tuned
flow-matching action expert \\
Observation inputs &
Two third-person images, one wrist image, language instruction, and
proprioceptive state \\
Image resolution & $224\times224$ pixels per view \\
Image augmentation & 93.3\% random crop and color jitter \\
Latent-code readout &
Q-Former-style readout from 16 latent plan tokens to a 16-dim code \\
Action output &
12-dimensional mobile-manipulation actions in a 16-step action chunk \\
Action decoder & Flow-matching action expert \\
VLM-KV condition dropout &
Training only: sample-level probability 0.10 of replacing the complete VLM
KV sequence with shared learned null tokens \\
\midrule
\rowcolor{PearlGroupGray}\multicolumn{2}{@{}l}{\textit{Latent refinement}} \\
Refinement rounds & $K=3$ \\
Plan update &
Fixed round-wise scalars with nominal update RMS of 15\%, 6\%, and 3\% \\
\midrule
\rowcolor{PearlGroupGray}\multicolumn{2}{@{}l}{\textit{Compute}} \\
Hardware & $4\times$ NVIDIA A100 40GB GPUs \\
\bottomrule
\end{tabularx}
\end{table}
\FloatBarrier

\section{Computational Overhead and Inference Efficiency}
\label{app:efficiency}

\subsection{Measurement Setup}

All measurements were collected on one NVIDIA A100 40GB GPU with batch size
one and PyTorch eager execution. The host VLM used BF16 autocast, and all
remaining modules used FP32. The $K=0$ configuration comprises one VLM forward
pass and one action-decoding pass without DINOv3 feature extraction or latent
refinement. For $K\geq1$, DINOv3 feature extraction is performed once; the
loop then computes the detached baseline prediction $\hat u_0$ and $K$ live
refinement predictions before the same single action decode. Thus $K$ denotes
refinement rounds, while the number of LaWM queries is $K+1$. The subgoal
renderer and reward model are excluded because neither is used during
deployment; CRG-PRL adds no deployment query.

We report policy-query latency, peak GPU memory, GFLOPs, and effective action
throughput, computed as $H\times1000/\mathrm{latency}$. Consistent with our
evaluation protocol, the complete predicted action chunk is executed without
temporal ensembling before the next policy query. Effective action throughput
is not the policy-query rate, and values from different action-chunk lengths
are therefore not directly comparable.

\subsection{Efficiency Results}

Tables~\ref{tab:eff-openvla} and~\ref{tab:eff-groot} report results for the
OpenVLA-7B / LIBERO ($H=8$) and GR00T N1.5 / RoboCasa ($H=16$, four Euler
steps) instantiations, respectively.

\begin{table}[!ht]
\caption{Inference efficiency on OpenVLA-7B / LIBERO.}
\label{tab:eff-openvla}
\centering
\small
\resizebox{\linewidth}{!}{
\begin{tabular}{@{}lrrrr@{}}
\toprule
Configuration &
Latency (ms) $\downarrow$ &
Throughput (Hz) $\uparrow$ &
Peak memory (GB) $\downarrow$ &
GFLOPs $\downarrow$ \\
\midrule
VLM + action head ($K=0$) & 156.6 & 51 & 15.4 & 6{,}487 \\
PearlVLA ($K=1$) & 176.3 & 45 & 16.3 & 6{,}663 \\
PearlVLA ($K=2$) & 187.5 & 43 & 16.3 & 6{,}770 \\
PearlVLA ($K=3$) & 198.7 & 40 & 16.3 & 6{,}878 \\
\bottomrule
\end{tabular}}
\end{table}

\begin{table}[!ht]
\caption{Inference efficiency on GR00T N1.5 / RoboCasa.}
\label{tab:eff-groot}
\centering
\small
\resizebox{\linewidth}{!}{
\begin{tabular}{@{}lrrrr@{}}
\toprule
Configuration &
Latency (ms) $\downarrow$ &
Throughput (Hz) $\uparrow$ &
Peak memory (GB) $\downarrow$ &
GFLOPs $\downarrow$ \\
\midrule
VLM + action expert ($K=0$) & 59.0 & 271 & 5.4 & 2{,}089 \\
PearlVLA ($K=1$) & 78.7 & 203 & 6.2 & 2{,}266 \\
PearlVLA ($K=2$) & 89.9 & 178 & 6.2 & 2{,}373 \\
PearlVLA ($K=3$) & 101.0 & 158 & 6.2 & 2{,}480 \\
\bottomrule
\end{tabular}}
\end{table}
\FloatBarrier

The three-round refinement loop adds about 42ms on both instantiations. The
first refinement step accounts for
about 20ms and 176 GFLOPs, covering one-time DINOv3 extraction, $\hat u_0$,
and the first live LaWM query; each later round adds one live query and about
11ms and 107 GFLOPs. Peak memory rises by 0.9GB for OpenVLA and 0.8GB for
GR00T, and does not grow with further rounds. With this overhead, LaWM-guided
refinement provides effective action throughputs of 40Hz on OpenVLA and 158Hz
on GR00T.

\section{Causal Refinement-Grouped Process-Reward RL}
\label{app:crg-prl}

CRG-PRL optimizes the refinement operator already learned by the supervised
PearlVLA policy. The reward is a proxy produced by a frozen LaWM, renderer, and
robot process reward model. Its central design follows one principle: edits should only be compared under the \emph{same} refinement state,
the \emph{same} round, and the \emph{same} reward pipeline.
This section defines the inner decision process, same-state local branching, post-edit process reward,
and critic-free objective.

\subsection{Inner MDP and Policy}
\label{sec:crg-policy}

For a fixed observation and instruction, let $\tilde z_0$ denote the initial
plan after the host-specific perturbation $\mathcal I_{\mathrm{host}}$, which
applies the OpenVLA perturbation with scale $\rho_{\mathrm{init}}=0.1$ and is
the identity for GR00T, and let $z_0$ denote the writable loop state
initialized from $\tilde z_0$.
The supervised loop initializes
\begin{equation}
\begin{aligned}
c_0&=C(\tilde z_0),\\
\hat u_0&=\sg(F_\omega(u_t,c_0)),
\qquad
c_1=\sg(c_0)+\operatorname{Composer}(z_0,\sg(c_0)).
\end{aligned}
\label{eq:crg-init}
\end{equation}
At round $k=1,\ldots,K$, it queries
$\hat u_k=F_\omega(u_t,c_k)$ and constructs the relative-imagination input and
frozen future summary
\begin{equation}
\tau_k =
\left[\hat u_k-u_t\,;\,\hat u_k-\hat u_0\right],
\qquad
h_k=\operatorname{FutureEncoder}(\tau_k),
\label{eq:crg-summary}
\end{equation}
where $\hat u_0$ is the shared first-prediction reference. To avoid overloading
the renderer parameter $\xi$, denote the inner state by
\begin{equation}
x_k=(z_{k-1},h_k,c_k,k).
\end{equation}
The round index makes the finite-horizon state explicit and selects the
prescribed write-back and exploration scales introduced below; the shared
RefineNet mean itself receives only $(z_{k-1},h_k,c_k)$.
RefineNet maps this state to a bottleneck mean
$\mu_\theta(x_k)\in\mathbb R^{N_p\times384}$, where $N_p=8$ for LIBERO and
$N_p=16$ for RoboCasa. CRG-PRL defines a fixed-variance Gaussian policy in
this bottleneck:
\begin{equation}
b_k\sim\pi_\theta^b(\cdot\mid x_k)
=\mathcal N\!\left(\mu_\theta(x_k),\sigma_k^2I\right),
\qquad
\delta_k=U_{\mathrm{up}}b_k,
\label{eq:crg-bottleneck-policy}
\end{equation}
where the frozen linear up-projection $U_{\mathrm{up}}$ maps the bottleneck to
$N_p\times4096$. The resulting transition is
\begin{equation}
z_k=z_{k-1}+w_k\delta_k,
\qquad
c_{k+1}=\sg(c_0)+\operatorname{Composer}(z_k,c_k).
\label{eq:crg-transition}
\end{equation}
Only the pre-expansion layers of RefineNet are updated during CRG-PRL;
FutureEncoder and $U_{\mathrm{up}}$ remain frozen. Deployment disables
exploration and uses $b_k=\mu_\theta(x_k)$.

The exploration scale retains one dimensionless hyperparameter across rounds.
Let $\mu_{\mathrm{SFT}}$ denote the frozen supervised RefineNet mean used as
the calibration reference, and define
\begin{equation}
\nu_k =
\operatorname{EMA}\!\left[
\operatorname{RMS}\!\left(
w_kU_{\mathrm{up}}\mu_{\mathrm{SFT}}(x_k)
\right)\right],
\qquad
g_{\mathrm{up}}=
\sqrt{\frac{\lVert U_{\mathrm{up}}\rVert_F^2}{4096}}.
\end{equation}
The round-specific scale is
\begin{equation}
\sigma_k=
\frac{\sigma_{\mathrm{norm}}\nu_k}
{w_kg_{\mathrm{up}}}.
\label{eq:crg-sigma}
\end{equation}
The statistics $\nu_k$ are estimated once from frozen SFT rollouts before
CRG-PRL, and the resulting $\sigma_k$ values remain fixed throughout all
updates.
Thus, in RMS, the realized write-back noise is approximately a fraction
$\sigma_{\mathrm{norm}}$ of the natural SFT edit on the calibration set in
every round; it is not a per-state equality. We use
$\gamma=0$, yielding a refreshed local-ranking surrogate at states along the
deterministic deployment trajectory rather than an objective over multi-round
returns.

\subsection{Same-State Local Branching}
\label{sec:crg-branching}

Each outer iteration collects fresh data under the current rollout snapshot
$\theta_{\mathrm{roll}}$. It first runs a deterministic base path:
\begin{equation}
\begin{aligned}
b_k^{\mathrm{base}}
&=\mu_{\theta_{\mathrm{roll}}}(x_k^{\mathrm{base}}),\\
z_k^{\mathrm{base}}
&=z_{k-1}^{\mathrm{base}}
+w_kU_{\mathrm{up}}b_k^{\mathrm{base}},\\
c_{k+1}^{\mathrm{base}}
&=\sg(c_0)+\operatorname{Composer}
\!\left(z_k^{\mathrm{base}},c_k^{\mathrm{base}}\right).
\end{aligned}
\end{equation}
At every base state, we draw $M=8$ one-round branches,
\begin{equation}
b_k^{(i)}
=\mu_{\theta_{\mathrm{roll}}}(x_k^{\mathrm{base}})
+\sigma_k\eta_k^{(i)},
\qquad
\eta_k^{(i)}\sim\mathcal N(0,I),
\quad i=1,\ldots,M.
\label{eq:crg-branches}
\end{equation}
All branches in a group share the observation features, latent plan, committed
code, first prediction, current imagined future, instruction,
and frozen reward pipeline; only $b_k^{(i)}$ varies. A branch survives for one
round and is used only to evaluate that edit. It is not propagated into the
next base state, which prevents later-round groups from mixing different
starting states. Dropout is disabled during collection, and shared reward-side
randomness is used within each group.

The buffer stores detached base states, sampled bottlenecks, and advantages.
It stores $b_k^{(i)}$ itself rather than reconstructing it from
$\eta_k^{(i)}$ during optimization; otherwise the sampled action and
current mean would share a gradient path and invalidate the score-function
estimator. With $B=8$, $K=3$, and $M=8$, each distributed update contains
$BKM=192$ conditional branches. The complete fresh buffer is used for one
backward pass and one optimizer update, and is then discarded.

\subsection{Causal Process Reward}
\label{sec:crg-reward}

The current imagined future $\hat u_k$ precedes the sampled bottleneck and is
therefore constant within a group. CRG-PRL instead rewards the consequence
predicted after the edit:
\begin{equation}
\begin{aligned}
\delta_k^{(i)}
&=U_{\mathrm{up}}b_k^{(i)},\\
z_k^{(i)}
&=z_{k-1}^{\mathrm{base}}+w_k\delta_k^{(i)},\\
c_{k+1}^{(i)}
&=\sg(c_0)+\operatorname{Composer}
\left(z_k^{(i)},c_k^{\mathrm{base}}\right),\\
\hat u_{k,1}^{\mathrm{rew},(i)}
&=F_\omega(u_t,c_{k+1}^{(i)}),\\
\hat u_{k,2}^{\mathrm{rew},(i)}
&=F_\omega(\hat u_{k,1}^{\mathrm{rew},(i)},c_{\mathrm{null}}).
\end{aligned}
\label{eq:crg-reward-chain}
\end{equation}
Here the hop index $j\in\{1,2\}$ corresponds to physical time
$t+j\Delta$. Starting from $v_t=E_{\mathrm{vae}}(o_t^{\mathrm{wm}})$, the
frozen renderer and VAE decoder produce the reward images recursively:
\begin{equation}
\begin{aligned}
\hat v_{k,0}^{\mathrm{rew},(i)}
&=v_t,\\
\hat v_{k,j}^{\mathrm{rew},(i)}
&=R_\xi\!\left(
\hat v_{k,j-1}^{\mathrm{rew},(i)},
\hat u_{k,j}^{\mathrm{rew},(i)}
\right),\\
\hat o_{k,j}^{\mathrm{rew},(i)}
&=D_{\mathrm{vae}}\!\left(\hat v_{k,j}^{\mathrm{rew},(i)}\right),
\qquad j=1,2.
\end{aligned}
\label{eq:crg-reward-render}
\end{equation}
Robometer~\citep{liang2026robometer} scores the resulting ordered
three-frame, dual-view sequence conditioned on the instruction:
\begin{equation}
\begin{aligned}
O_k^{(i)}
&=\left[
o_t^{\mathrm{wm}},
\hat o_{k,1}^{\mathrm{rew},(i)},
\hat o_{k,2}^{\mathrm{rew},(i)}
\right],\\
\left(p_{k,i}^{\mathrm{prog}},p_{k,i}^{\mathrm{succ}}\right)
&=\operatorname{Robometer}_{\mathrm{final}}\!\left(O_k^{(i)},\ell\right),\\
r_k^{(i)}
&=p_{k,i}^{\mathrm{prog}}+0.2\,p_{k,i}^{\mathrm{succ}}.
\end{aligned}
\label{eq:crg-reward}
\end{equation}
For LIBERO, this path uses the third-person and wrist views; for RoboCasa, it
uses the external-left and wrist views. The second RoboCasa third-person view
is consumed only by the host VLM. For the final edit, LaWM is queried with
$c_{K+1}$ only by the training-time reward branch; $c_{K+1}$ also remains the
terminal SFT target, while the host action is decoded from $z_K$. Reward
rollout, rendering, and scoring run without gradients.

The term ``causal'' refers to this controlled model-internal intervention:
the pre-edit state and frozen proxy pipeline are held fixed while only the
bottleneck edit changes. It does not assert that the proxy reward establishes
physical causality in the real environment. We therefore monitor latent-code
support, reward spread, renderer stability, and final-action sensitivity to
detect proxy exploitation.

\subsection{Group-Relative Advantage}
\label{sec:crg-advantage}

For each sample-round group, rewards are standardized using the population
standard deviation:
\begin{equation}
\bar r_k=\frac{1}{M}\sum_{i=1}^{M}r_k^{(i)},
\qquad
s_k^r=
\sqrt{\frac{1}{M}\sum_{i=1}^{M}
\left(r_k^{(i)}-\bar r_k\right)^2},
\qquad
A_k^{(i)}
=\sg\!\left[
\frac{r_k^{(i)}-\bar r_k}{s_k^r+10^{-6}}
\right].
\label{eq:crg-advantage}
\end{equation}
This same-state baseline centers additive reward offsets specific to the state
and round without a learned critic. No explicit skip threshold is used: a
group with identical rewards has zero advantages, while $10^{-6}$ stabilizes
near-zero spread.
Because both the mean and standard deviation are estimated from the same finite
group, this is a biased group-normalized surrogate rather than an unbiased
gradient estimator of raw expected reward. Its role is local ranking under a
matched refinement state.

\subsection{Group-Relative Policy Objective}
\label{sec:crg-policy-objective}

Let $B$ denote the global task-sample batch size and let $d_b=384N_p$.
Because the covariance is fixed, the bottleneck log-density is
\begin{equation}
\log\pi_\theta^b(b_k^{(i)}\mid x_k)
=
-\frac{\left\|b_k^{(i)}-\mu_\theta(x_k)\right\|_F^2}
{2\sigma_k^2}
-d_b\log\sigma_k
-\frac{d_b}{2}\log(2\pi).
\end{equation}
Restoring the task-sample index $n$, the fresh-buffer score-function objective
is
\begin{equation}
\mathcal L_{\mathrm{PG}}
=-\frac{1}{BKM}
\sum_{n=1}^{B}\sum_{k=1}^{K}\sum_{i=1}^{M}
\sg\!\left(A_{nk}^{(i)}\right)
\log\pi_\theta^b\!\left(
\sg(b_{nk}^{(i)})\mid\sg(x_{nk}^{\mathrm{base}})
\right).
\label{eq:crg-pg}
\end{equation}
Event dimensions are summed to form one joint log-probability for each branch.
The sampled bottleneck, base state, and advantage are detached; gradients flow
only through the current policy mean. Because each buffer is collected from the
current policy, used for one optimizer update, and then discarded, no
importance ratio, clipping, replay, or value loss is required. This objective
uses the group-relative credit assignment of GRPO~\citep{shao2024deepseekmath}
without its completion-level sampling unit.
The bottleneck mean is cast to FP32, Gaussian noise is sampled in FP32, and the
resulting $b_k^{(i)}$ is stored in FP32. The bottleneck difference,
squared-norm reduction, and Gaussian log-probability are likewise evaluated in
FP32.

\subsection{SFT Proximity, Action Anchor, and Total Objective}
\label{sec:crg-guardrails}

Two guardrails keep the optimized refiner close to supervised behavior. First,
the deterministic current-policy and SFT plan edits at a saved base state are
\begin{equation}
\Delta z_k^\theta
=w_kU_{\mathrm{up}}\mu_\theta(x_k),
\qquad
\Delta z_k^{\mathrm{SFT}}
=w_kU_{\mathrm{up}}\mu_{\mathrm{SFT}}(x_k).
\end{equation}
We penalize their difference:
\begin{equation}
\mathcal L_{\mathrm{prox}}
=\frac{1}{K}\sum_{k=1}^{K}
\mathbb E_{x_k}
\left[
\frac{
\operatorname{RMS}^2
\left(\Delta z_k^\theta-\Delta z_k^{\mathrm{SFT}}\right)}
{\nu_k^2}
\right].
\label{eq:crg-prox}
\end{equation}
This term is evaluated on saved deterministic base states and requires no new
LaWM rollout. Policy-to-SFT KL remains a diagnostic rather than a loss.

Second, a fresh deterministic path is anchored to the supervised host target:
\begin{equation}
\mathcal L_{\mathrm{act}}
=
\ell_{\mathrm{host}}
\left(
\mathcal D_{\mathrm{host}}(z_K^\theta;\chi_t),
y_{\mathrm{host}}^\star
\right).
\label{eq:crg-action-anchor}
\end{equation}
For OpenVLA, $\ell_{\mathrm{host}}$ is the unscaled per-element mean action
regression loss; for GR00T, it is the unscaled per-element mean
flow-matching-velocity loss. The host decoder remains frozen, but gradients
propagate through it to the refined plan. Removing the host-specific
supervised multipliers of 25 and 10 prevents these implementation scales from
being folded into the shared action-anchor weight.
The GR00T anchor otherwise retains the host-native flow-matching construction;
VLM-KV condition dropout is disabled throughout CRG-PRL.

On the fresh deterministic anchor path, frozen $U_{\mathrm{up}}$, Composer,
LaWM, FutureEncoder, and the host decoder retain their input Jacobians, so the
action loss can reach earlier refinement rounds. Stop-gradient is applied only
to $c_0$ and $\hat u_0$, the complete reward branch, and the stored rollout
states, bottleneck samples, and advantages.

The total objective is
\begin{equation}
\mathcal L_{\mathrm{CRG}}
=
\mathcal L_{\mathrm{PG}}
+0.01\,\mathcal L_{\mathrm{prox}}
+0.1\,\mathcal L_{\mathrm{act}}.
\label{eq:crg-total}
\end{equation}
The three terms are normalized independently: $\mathcal L_{\mathrm{PG}}$ over
$BKM$ branches, $\mathcal L_{\mathrm{prox}}$ over $BK$ base states, and
$\mathcal L_{\mathrm{act}}$ over $B$ task samples and all action coordinates.
In distributed training each GPU uses the corresponding local denominators
$B_{\mathrm{local}}KM=48$, $B_{\mathrm{local}}K=6$, and
$B_{\mathrm{local}}=2$; DDP then averages the four local gradients. This is
equivalent to the stated global means and avoids an additional factor-of-four
shrinkage. Consequently, the loss weights do not change with the number of
branches, rounds, or GPUs.
Fixed covariance makes the entropy constant. Explicit SFT-KL and entropy terms
are therefore omitted; policy-to-SFT KL is retained as a diagnostic.

\subsection{Relation to Group-Relative LLM RL}

CRG-PRL borrows group-relative, critic-free credit assignment from GRPO and
GSPO~\citep{shao2024deepseekmath,zheng2025group}, but changes the
decision and credit-assignment units. GRPO and GSPO compare complete
prompt-conditioned completions or sequences; CRG-PRL compares one-round
continuous bottleneck edits at a fixed refinement state. WMPO optimizes action
trajectories inside a learned pixel world model and discards that model at
deployment~\citep{zhu2025wmpo}, while VLA-RFT uses a frozen video world model
to provide verified training rewards~\citep{vla-rft}. CRG-PRL instead
optimizes an edit operator inside the deployed refinement loop. The renderer
and Robometer are training-only, whereas LaWM remains part of policy inference.

\clearpage
\subsection{Implementation Details}

Table~\ref{tab:crg-implementation} summarizes the CRG-PRL configuration.
Branches from the same round are evaluated in parallel and are not counted as
additional task samples.
Reward-side branching, rendering, and scoring are used only during training;
deployment follows one deterministic three-round refinement path.

\begin{table}[!ht]
\caption{Core settings of CRG-PRL.}
\label{tab:crg-implementation}
\centering
\small
\begin{tabularx}{\linewidth}{@{}LY@{}}
\toprule
Setting & Value \\
\midrule
Refinement and branches &
$K=3$ rounds; $M=8$ same-state branches per round \\
Bottleneck policy &
LIBERO: $8\times384$; RoboCasa: $16\times384$; fixed isotropic covariance \\
Deterministic expansion &
Frozen linear up-projection to $8\times4096$ or $16\times4096$ \\
Exploration &
$\sigma_{\mathrm{norm}}=0.06$, converted to round-specific $\sigma_k$ by
Eq.~\ref{eq:crg-sigma}, pre-calibrated once and then fixed \\
Updated parameters &
RefineNet except the final up-projection; learning rate $1\times10^{-5}$ \\
Frozen parameters &
Host VLM and interfaces, $C$, FutureEncoder, up-projection, Composer, $w_k$,
LaWM, renderer, Robometer, host policy's action interface, and SFT RefineNet reference \\
Reward &
Dual-view $[o_t^{\mathrm{wm}},
\hat o_{k,1}^{\mathrm{rew}},\hat o_{k,2}^{\mathrm{rew}}]$; final-frame progress
$+\,0.2\times$ success from frozen Robometer \\
Loss weights &
$\lambda_{\mathrm{prox}}=0.01$ and $\lambda_{\mathrm{act}}=0.1$; no value,
explicit SFT-KL, or entropy loss \\
Group advantage &
Population standard deviation plus $10^{-6}$; no explicit group skipping \\
Policy update &
Fresh-buffer group-relative policy gradient with $\gamma=0$; no replay,
importance ratio, or clipping \\
Optimization and compute &
2000 outer iterations and optimizer updates; two task samples per GPU;
$4\times$ NVIDIA A100 40GB GPUs; global rollout batch size 8 \\
Diagnostics &
Raw reward, per-round group standard deviation, policy-to-SFT KL, joint
log-probability, policy-gradient norm, realized edit RMS, code-support
distance, zero-advantage group rate, and the correlation between terminal
$c_{K+1}$ reward and host-action quality \\
\bottomrule
\end{tabularx}
\end{table}
\FloatBarrier

\end{document}